\documentclass[11pt,a4paper]{article}
\usepackage[hyperref]{acl2020}
\usepackage{times}
\usepackage{latexsym}

\usepackage{microtype}

\aclfinalcopy 
\def\aclpaperid{2232} 

\usepackage {spverbatim}

\usepackage{CJKutf8}

\title{A Supervised Word Alignment Method based on Cross-Language Span
  Prediction using Multilingual BERT}

\author{Masaaki Nagata \quad Chousa Katsuki \quad
  Masaaki Nishino \\
  NTT Communication Science Laboratories, NTT Corporation \\
  2-4, Hikaridai, Seika-cho, Soraku-gun, Kyoto, 619-0237, Japan \\
  \texttt{\{masaaki.nagata.et,katsuki.chousa.bg,masaaki.nishino.uh\}@hco.ntt.co.jp} \\
}

\date{}

\begin{document}
\maketitle
\begin{abstract}
  We present a novel supervised word alignment method based on
  cross-language span prediction. We first formalize a word alignment
  problem as a collection of independent predictions from a token in
  the source sentence to a span in the target sentence. As this is
  equivalent to a SQuAD v2.0 style question answering task, we then
  solve this problem by using multilingual BERT, which is fine-tuned
  on a manually created gold word alignment data.  We greatly improved
  the word alignment accuracy by adding the context of the token to
  the question. In the experiments using five word alignment datasets
  among Chinese, Japanese, German, Romanian, French, and English, we
  show that the proposed method significantly outperformed previous
  supervised and unsupervised word alignment methods without using any
  bitexts for pretraining. For example, we achieved an F1 score of 86.7
  for the Chinese-English data, which is 13.3 points higher than the previous
  state-of-the-art supervised methods.

\end{abstract}

\section{Introduction}

Over the past six years, machine translation accuracy has greatly
improved by using neural networks
\cite{Cho_etal_EMNLP2014,Sutskever_etal_NIPS2014,Bahdanau_etal_ICLR2015,Luong_etal_EMNLP2015,Vaswani_etal_NIPS2017}.
Unfortunately, word alignment accuracy has not greatly improved nealy
20 years. Word alignment tools, which were developed during the age of statistical
machine translation \cite{Brown_etal_CL1993,Koehn_etal_ACL2007} such
as GIZA++ \cite{Och_Ney_CL2003}, MGIZA \cite{Gao_Vogel_ACLWS2008} and
FastAlign \cite{Dyer_etal_NAACL2013}, remain widely used.

This situation is unfortunate because word alignment could be used for
many downstream tasks include projecting linguistic annotation
\cite{Yarowsky_etal_HLT2001}, projecting XML markups
\cite{Hashimoto_etal_WMT2019}, and enforcing terminology constraints
(pre-specified translation) \cite{Song_etal_NAACL2019}. We could also
use it for user interfaces of post-editing to detect problems such as
under-translation \cite{Tu_etal_ACL2016}.

Previous works used neural networks for word alignment
\cite{Yang_etal_ACL2013,Tamura_etal_ACL2014,Legrand_etal_WMT2016}, but
their accuracies were at most comparable to that of GIZA++.  Recent
works \cite{Zenkel_etal_arXiv2019,Garg_etal_EMNLP2019} try to make the
attention of the Transformer to be close to word alignment, and
\citet{Garg_etal_EMNLP2019} achieved slightly better word
alignment accuracy than that of GIZA++ when alignments obtained from
GIZA++ is used for supervision.

In contrast to these unsupervised approaches,
\citet{Stengel-Eskin_etal_EMNLP2019} proposed a supervised word
alignment method that significantly outperforms FastAlign (11-27 F1
points) using a small number of gold word alignments (1.7K-5K
sentences). However, both \cite{Garg_etal_EMNLP2019} and
\cite{Stengel-Eskin_etal_EMNLP2019} require more than a million
parallel sentences to pretrain their transformer-based model.

In this paper, we present a novel supervised word alignment method
that does not require parallel sentences for pretraining and can be
trained an even smaller number of gold word alignments (150-300
sentences). It formalizes word alignment as a collection of
SQuAD-style span prediction problems \cite{Rajpurkar_etal_EMNLP2016}
and uses multilingual BERT \cite{Devlin_etal_NAACL2019} to solve
them. We show that, by experiment, the proposed model significantly
outperforms both \cite{Garg_etal_EMNLP2019} and
\cite{Stengel-Eskin_etal_EMNLP2019} in word alignment accuracy.

\citet{McCann_etal_arXiv2018} formalizes a variety of natural language
tasks as a question answering problem.  Multilingual BERT can be used
for a variety of (zero-shot) cross-language applications such as named
entity recognition
\cite{Pires_elal_ACL2019,Wu_Dredze_EMNLP2019}. However, to the best of
our knowledge, ours is the first work that formalizes word alignment
as question answering and adopts multilingual BERT for word alignment.

\section{Proposed Method}

\subsection{Word Alignment as Question Answering}

\begin{figure*}[tb]
  \centering
  \small
\begin{CJK}{UTF8}{min}
\begin{spverbatim}
足利 義満 （ あしかが よしみつ ） は 室町 幕府 の 第 3 代 征夷 大 将軍 （ 在位 1368 年 - 1394 年 ） で あ る 。
yoshimitsu ashikaga was the 3rd seii taishogun of the muromachi shogunate and reigned from 1368 to1394 .
0-1 1-0 3-1 4-0 7-9 8-10 9-7 10-3 11-4 12-4 13-5 14-6 15-6 17-12 18-14 19-14 21-15 22-15 24-2 25-2 26-2 27-16
足利義満（あしかがよしみつ）は室町幕府の第3代征夷大将軍（在位1368年-1394年）である。
Yoshimitsu ASHIKAGA was the 3rd Seii Taishogun of the Muromachi Shogunate and reigned from 1368 to1394.
\end{spverbatim}
\end{CJK}
  \caption{Example of word alignment data between Japanese and English}
  \label{fig:input_example}
\end{figure*}

Fig.~\ref{fig:input_example} shows an example of word alignment
data. It consists of a token sequence of the L1 language (Japanese), a
token sequence of the L2 language (English), a sequence of aligned token
pairs, the original L1 sentence, and the original L2 sentence.  For example,
the first item of the third line, ``0-1,'' represents the first token,
\begin{CJK}{UTF8}{min}``足利,''\end{CJK} of the L1 sentence is aligned
to the second token ``ashikaga'' of the L2 sentence. The index of the
tokens starts from zero.

\begin{figure*}[tb]
  \centering
  \small
\begin{CJK}{UTF8}{min}
\begin{spverbatim}
context: "足利義満（あしかがよしみつ）は室町幕府の第3代征夷大将軍（在位1368年-1394年）である。"
question: "was"
answer: "である"
\end{spverbatim}
\end{CJK}
  \caption{Example of an English-to-Japanese query without source context}
  \label{fig:squad_example_no_context}
\end{figure*}

Fig.~\ref{fig:squad_example_no_context} shows an example in which the
aligned tokens are converted to the SQuAD-style span prediction.  Here
the L1 (Japanese) sentence is given as the context. A token in the L2
(English) sentence ``was'' is given as the question whose answer is
span \begin{CJK}{UTF8}{min}``である''\end{CJK} in the L1 sentence. It
corresponds to the three aligned token pairs ``24-2 25-2 26-2'' in the
third line of Fig.~\ref{fig:input_example}.

As shown in the above examples, we can convert word alignments for a
sentence to a set of queries from a token in the L1 sentence to a span
in the L2 sentence, and a set of queries from a token in the L2
sentence to a span in the L1 sentence. If a token is aligned to
multiple spans, we treat the question has multiple answers. If a token
has no alignment, we treat there are no answers to the question.

In this paper, we call the language of the question the source language
and the language of the context (and the answer) the target language. In
Fig.~\ref{fig:squad_example_no_context}, the source language is
English and the target language is Japanese. We call the query
English-to-Japanese query.

\begin{figure*}[htb]
  \centering
  \small
\begin{CJK}{UTF8}{min}
\begin{spverbatim}
context: "足利義満（あしかがよしみつ）は室町幕府の第3代征夷大将軍（在位1368年-1394年）である。"
question: "Yoshimitsu ASHIKAGA ¶ was ¶ the 3rd"
answer: "である",
\end{spverbatim}
\end{CJK}
  \caption{Example of an English-to-Japanese query with a short source context}
  \label{fig:squad_example_short_context}
\end{figure*}

Suppose the question is such a high-frequency word as ``of'', which
might be found many times in the source sentence, we could easily have
difficulty in finding the corresponding span in the target sentence
without the source token's context.

Fig.~\ref{fig:squad_example_short_context} shows an example of a
question with a short context of the source token.  The two
preceding words ``Yoshimitsu ASHIKAGA'' and two following words ``the
3rd'' in the source sentence are attached to the source token ``was''
with `¶' (pilcrow: paragraph mark) as a boundary marker\footnote{ We
  used `¶' as a boundary marker because it belongs to Unicode
  character category ``punctuation'' and it is included in the multilingual BERT vocabulary. It looks like `$|$' and it rarely
  appears in ordinary text.}.

\begin{figure*}[htb]
  \centering
  \small
\begin{CJK}{UTF8}{min}
\begin{spverbatim}
{
  "version": "v2.0",
  "data": [
    {
      "paragraphs": [
        {
          "context": "Yoshimitsu ASHIKAGA was the 3rd Seii Taishogun of the Muromachi Shogunate and reigned from 1368 to1394.",
          "qas": [
            ...
            {
              "id": "kftt_devtest_0_f_1_0",
              "question": "足利 ¶ 義満 ¶ （あしかがよしみつ）は室町幕府の第3代征夷大将軍（在位1368年-1394年）である。",
              "answers": [
                {
                  "text": "Yoshimitsu",
                  "answer_start": 0
                }
              ],
              "is_impossible": false
            },
            ...
\end{spverbatim}
\end{CJK}          
  \caption{SQuAD v2.0 json format of a Japanese-to-English query with
    full source context.}
  \label{fig:squad_example}
\end{figure*}

As we show in the experiment, the longer the context is, the
better. We decided to use the whole source sentence as a
context. Since there are many null alignments in word alignment, we
adopted SQuAD v2.0 format \cite{Rajpurkar_elal_ACL2018}, which
supports cases when there are no answer spans to the question in the
given context.  Fig.~\ref{fig:squad_example} shows an example of
word alignment data in the SQuAD v2.0 format.

In Fig.~\ref{fig:squad_example}, both the question and the context are
taken from the original sentences, not the tokenized sequences.  The
feature ``answer\_start'' is an index to the context's character
position. The tokens in the word alignment data are used only for
generating questions.

The feature ``is\_impossible'' is true if there are no answers to the
question. It is an extension from SQuAD v1.1 to v2.0. It is essential
to model null alignment (no answers) explicitly. In a preliminary
experiment, we used the SQuAD v1.1 format and got unsatisfactory
results.


\subsection{Cross-Language Span Prediction using Multilingual BERT}

We defined our cross-language span prediction task as follows: Suppose
we have a source sentence with N characters
$F = \{ f_{1}, f_{2}, \dots, f_{N} \}$, and a target sentence with M
characters $E = \{ e_{1}, e_{2}, \dots, e_{M} \}$.  Given a source
token $Q(i,j) = \{ f_{i}, f_{i+1}, \dots, f_{j-1} \}$ that spans
$(i, j)$ in the source sentence $F$, the task is to extract target
span $R(k,l) = \{ e_{k}, e_{k+1}, \dots, e_{l-1} \}$ that spans
$(k, l)$ in the target sentence $E$.

We applied multilingual BERT \cite{Devlin_etal_NAACL2019} to this
task. Although it is designed for such monolingual language
understanding tasks as question answering and natural language
inference, it also works surprisingly well for the cross-language span
prediction task.

We used the model for SQuAD v2.0 described in
\citet{Devlin_etal_NAACL2019}. It adds two independent output layers to
pretrained (multilingual) BERT to predict the start and end positions
in the context.  Suppose $p_{start}$ and $p_{end}$ are the
probabilities that each position in the target sentence is the start
and end positions of the answer span to the source token. We define
the score of a span $\omega$ as the product of its start and end
position probabilities and select the span $(\hat{k}, \hat{l})$ which
maximizes $\omega$ as the best answer span:

\begin{equation}
  \label{eq:span_score}
  \omega_{ijkl}^{F \rightarrow E}
  = p_{start} (k | E, F, i, j) \cdot p_{end} (l | E, F, i, j)
\end{equation}
\begin{equation}
  \label{eq:span_argmax}
  ( \hat{k}, \hat{l} ) = \arg \max_{(k,l): 1 \leq k \leq l \leq M}
  \omega_{ijkl}^{F \rightarrow E}
\end{equation}

In the SQuAD model of BERT, first, the question and the context is
concatenated to generate a sequence ``[CLS] question [SEP] context
[SEP]'' as input, where `[CLS]' and `[SEP]' are classification token
and separator token, respectively. Then, the start and end positions
are predicted as indexes to the sequence. In the SQuAD v2.0 model, the
start and end positions are the indexes to the [CLS] token if there
are no answers.

Unfortunately, since the original implementation of the BERT SQuAD
model only outputs an answer string, we extended it to output the
answer's start and end positions. Inside BERT, the input sequence is
first tokenized by WordPiece. It then splits the CJK charactes into a
sequence of a single character. As the start and end positions are
indexes to the BERT subtokens, we converted them to the character
indexes to the context, considering that the offset for the context
tokens is the length of question tokens plus two ([CLS] and [SEP]).

\subsection{Symmetrization of Word Alignments}

Since the proposed span prediction model predicts a target span for a
source token, it is asymmetric as the IBM model
\citet{Brown_etal_CL1993}. To support a one-to-many-span
alignment, and to make the prediction of spans
more reliable, we designed a simple heuristics to symmetrize the
span predictions of different directions.

Symmetrizing IBM model alignments was first proposed by
\cite{Och_Ney_CL2003}. One of the most popular Statistical Machine
Translation Toolkit, Moses \cite{Koehn_etal_ACL2007}, supports a
variety of symmetrization heuristics, such as intersection and union,
in which grow-diag-final-and is the default. The intersection of the
two alignments yields an alignment that consists of only one-to-one
alignments with higher precision and lower recall than either one
separately. The union of the two alignments yields higher recall and
lower precision.

As a symmetrization method, for an alignment that consists of a pair
of a L1 token and a L2 token, we average the probabilities of the best
spans for each token for each direction. We treat a token as aligned if
it is completely included in the predicted span. We then extract the
alignments with the average probabilities that exceed a threshold
$\theta$.

Let p and q be start and end character indexes to sentence F, and let
r and s be the start and end character indexes to sentence E.  Let
$\omega_{pqrs}^{F \rightarrow E}$ be the probability that a token that
starts p and ends q in F, predicts a span that starts r and ends s in
E.  Let $\omega_{pqrs}^{E \rightarrow F}$ be the probability that a
token that starts r and ends s, predicts a span that starts p and end
q in F.  We define the probability of alignment $a_{ijkl}$, which
represents that a token stats i and ends j in F, is aligned to a token
that starts k and ends l in E:

\begin{equation}
  \label{eq:symmetrization}
  P(a_{ijkl}) = 1/2 (
  \sum_{r \leq k \leq l \leq s} \omega_{ijrs}^{F \rightarrow E} +
  \sum_{p \leq i \leq j \leq q} \omega_{pqkl}^{E \rightarrow F}
  )
\end{equation}

Here, we set the threshold to 0.4, which means that, if the sum of the
probabilities of both directions exceeds 0.8, the alignment is
selected. For example, if the probability of Ja-to-En is 0.9, the
alignment selected even if the probability of En-to-Ja is 0. If the
probability of Ja-to-En is 0.5 and that of En-to-Ja is 0.4, the
alignment is also selected.

We determined a threshold of 0.4 in a preliminary experiment in which
we divided the Japanese-English training data into two halves for
training and test sets. We used this threshold for all the experiments
reported in this paper. Although the span prediction of each direction
is made independently, we did not normalize the scores before
averaging because both directions are trained in one model.

\citet{Zenkel_etal_arXiv2019} subtokenized words by Byte Pair Encoding
\cite{Sennrich_etal_ACL2016} and applied GIZA++ over the
subtokens. They then considered two words to be aligned if any
subtokens are aligned.  In Eq.~\ref{eq:symmetrization}, we could treat
two tokens as aligned if a token and the predicted span overlap. We
could use not only the best span but also the nbest spans for the
summation in Eq.~\ref{eq:symmetrization}. These are our future works.

\section{Experiments}

\subsection{Data}

\begin{table}
  \begin{tabular}{l|rrr}
    Language & Train & Test & Reserve \\
    \hline
    Zh-En    & 4,879 &  610 &  610 \\
    Ja-En    &   653 &  357 &  225 \\
    De-En    &   300 &  208 &    0 \\
    Ro-En    &   150 &   98 &    0 \\
    En-Fr    &   300 &  147 &    0 \\
  \end{tabular}
  \centering
  \caption{Number of gold alignment sentences and their train/test splits.}
  \label{tab:datasets}
\end{table}

Table~\ref{tab:datasets} shows the number of training and test
sentences of the five gold word alignment datasets used in our
experiments: Chinese-English (Zh-En), Japanese-English (Ja-En),
German-English (De-En), Romanian-English (Ro-En), and English-French
(En-Fr).

\citet{Stengel-Eskin_etal_EMNLP2019} used the Zh-En dataset and
\citet{Garg_etal_EMNLP2019} used the De-En, Ro-En, and En-Fr
datasets. We added a Ja-En dataset because Japanese is one of the most
distant languages from
English\footnote{\citet{Stengel-Eskin_etal_EMNLP2019} also used an
  Arabic-English (Ar-En) dataset. We did not use it here due to time
  constraints}.

The Zh-En data were obtained from GALE Chinese-English Parallel
Aligned Treebank \cite{Li_etal_LDC2015}, which consists of
broadcasting news, news wire, and web data.  To make the experiment's
condition as close as possible to \cite{Stengel-Eskin_etal_EMNLP2019},
we used Chinese character-tokenized bitexts, cleaned them (by removing
mismatched bitexts, time stamps, etc.), and randomly split them as
follows: 80\% for training, 10\% for testing, and 10\% for future
reserves.

The Japanese-English data were obtained from the KFTT word alignment
data \cite{Neubig_KFTT2011}.  KFTT (Kyoto Free Translation Task)
\footnote{http://www.phontron.com/kftt/index.html} was made by
manually translating Japanese Wikipedia pages about Kyoto into
English.  It is one of the most popular Japanese-English translation
benchmarks. It consists of 440k training sentences, 1166 development
sentences, and 1160 test sentences.  The KFTT word alignment data were
made by manually word aligning a part of the dev and the test
sets. The aligned dev set has eight files and the aligned test set has
seven files. We used all eight dev set files for training, and four
test set files for testing, and three other files for future reserves.

De-En, Ro-En, and En-Fr data are the same ones described in
\cite{Zenkel_etal_arXiv2019}. They provide pre-processing and scoring
scripts\footnote{https://github.com/lilt/alignment-scripts}.
\citet{Garg_etal_EMNLP2019} also used these three datasets for their
experiments.  The De-En data were originally provided by
\cite{Vilar_etal_IWSLT2006}\footnote{https://www-i6.informatik.rwth-aachen.de/goldAlignment/}.
Ro-En and En-Fr data were used in the shared task of the
HLT-NAACL-2003 workshop on Building and Using Parallel Texts
\cite{Mihalcea_Pedersen_NAACLWS2003}\footnote{http://web.eecs.umich.edu/~mihalcea/wpt/index.html}. The
En-Fr data were originally provided by \cite{Och_Ney_ACL2000}.  The
numbers of test sentences in the De-En, Ro-En, and En-Fr datasets are
508, 248, and 447, respectively. In De-En and En-Fr, we used 300
sentences for training. In Ro-En, we used 150 sentences for
training. The other sentences were used for testing.

\subsection{Implementation Details}

We used BERT-Base, Multilingual Cased (104 languages, 12-layer,
768-hidden, 12-heads, 110M parameters, November 23rd, 2018) in our
experiments\footnote{https://github.com/google-research/bert}. We
basically used the script for SQuAD as it is. The following are the
parameters: train\_batch\_size = 12, learning\_rate = 3e-5,
num\_train\_epochs = 2, max\_seq\_length = 384, max\_query\_length =
160, and max\_answer\_length = 15.

In \citet{Devlin_etal_NAACL2019}, they used the following threshold
for the squad-2.0 model,

\begin{equation}
  \label{eq:non_null_threshold}
  \hat{s_{ij}} > s_{null} + \tau
\end{equation}

Here, if the difference of the score of best non-null span
$\hat{s_{ij}}$ and that of null (no-answer) span $s_{null}$ is exceeds
threshold $\tau$, a non-null span is predicted.  The default value of
$\tau = 0.0$, and optimal threshold is decided by the development
set. We used the default value because we assumed the score of a null
alignment is appropriately estimated as there are many null alignments
in the training data.

We used two NVIDIA TESLA V100 (16GB) for our experiments. Most of them
were performed in one GPU, but we sometimes faced out of memory errors
with just one GPU. If we set the training batch size to 6, the
experiments could be performed in NVIDIA GEFORCE RTX 2080 Ti (11GB)
with no significant differences in accuracy. It takes about 30 minutes
for an epoch for Ja-En data (653 sentences) to fine-tune.

\subsection{Measures for Word Alignment Quality}

We evaluated the quality of word alignment using F1 score that
assigns equal weights to precision (P) and recall (R):

\begin{equation}
  \label{eq:F1}
  F_{1} = 2 \times P \times R / (P + R)
\end{equation}

We also used alignment error rate (AER) \cite{Och_Ney_CL2003} if
necessary because some previous works only reported it. Let the
quality of alignment $A$ be measured against a gold word alignment
that contains sure ($S$) and possible($P$) alignments.  Precision,
recall, and AER are defined as follows:

\begin{equation}
  \label{eq:precision_AER}
  Precision(A,P) = \frac{|P \cap A|}{|A|}
\end{equation}

\begin{equation}
  \label{eq:recall_AER}
  Recall(A,S) = \frac{|S \cap A|}{|S|}
\end{equation}

\begin{equation}
  \label{eq:AER}
  AER(S,P,A) = 1 - \frac{|S \cap A| + |P \cap A|}{|S| + |A|}
\end{equation}

As \citet{Fraser_Marcu_CL2007} pointed out, ``AER is broken in a way
that favors precision''. It should be used sparingly.  In previous
works, \citet{Stengel-Eskin_etal_EMNLP2019} uses precision, recall,
and F1, while \citet{Garg_etal_EMNLP2019} and
\citet{Zenkel_etal_arXiv2019} used the precision, recall, and AER
based on \cite{Och_Ney_CL2003}. If no distinction exsits between sure
and possible alignments, the two definitions of precision and recall
agree. Among the five datasets we used, De-En and En-FR makes a
distinction between sure and possible alignments.

\subsection{Results}

\begin{table*}[tb]
  \begin{tabular}{l|l|rr|r|r}
    Test set & Method & P & R & F1 & AER  \\
    \hline
    \hline
    Zh-En
             & FastAlign \cite{Stengel-Eskin_etal_EMNLP2019}
                      & 80.5 & 50.5 & 62.0 & - \\
             & DiscAlign \cite{Stengel-Eskin_etal_EMNLP2019}
                      & 72.9 &  74.0 & 73.4 & - \\
             & Our method & 84.4 & 89.2 & \textbf{86.7} & 13.3 \\
    \hline
    Ja-En & Giza++ \cite{Neubig_KFTT2011}
                      & 59.5 & 55.6 & 57.6 & 42.4 \\
             & Our method & 77.3 & 78.0 & \textbf{77.6} & \textbf{22.4} \\
    \hline
    \hline
    De-En & MGIZA (BPE, Grow-Diag-Final) \cite{Zenkel_etal_arXiv2019}
                      & 91.3  &  70.2  & - & 20.6 \\
             & Multi-task + GIZA++ supervised \cite{Garg_etal_EMNLP2019}
                      & - & - & - & 16.0 \\
             & Our method (sure + possible) & 89.9 & 81.7 & 85.6 & 14.4 \\
             & (based on \cite{Och_Ney_CL2003})
                      & 89.9 & 87.3 & - & \textbf{11.4} \\
    \hline
    Ro-En & MGIZA (BPE, Grow-Diag-Final)  \cite{Zenkel_etal_arXiv2019}
                      & 90.9 & 61.8 & - & 26.4 \\
             & Multi-task + GIZA++ supervised \cite{Garg_etal_EMNLP2019}
                      & - & - & - & 23.1 \\
             & Our method & 90.4 & 85.3 & 86.7 & \textbf{12.2} \\
    \hline
    En-Fr & MGIZA (BPE, Grow-Diag) \cite{Zenkel_etal_arXiv2019}
                      & 97.5 & 89.7 & - & 5.9 \\
             & Multi-task + GIZA++ supervised \cite{Garg_etal_EMNLP2019}
                      & - & - & - & 4.6 \\
             & Our method (sure + possible) & 88.6 & 53.4 & 66.6 & 33.3 \\
             & (based on \cite{Och_Ney_CL2003})
                      & 88.6 & 96.7 & - & 9.4 \\
             & Our method (sure only) & 86.2 & 70.8 & 77.8 & 22.2 \\
             & (based on \cite{Och_Ney_CL2003})
                      & 97.7 & 93.9 & - & \textbf{4.0} \\
  \end{tabular}
  \centering
  \caption{Best-effort comparison of proposed method with previous
    works}
  \label{tab:comparison}
\end{table*}

Table~\ref{tab:comparison} compares the proposed method with previous
works.  In all five datasets, our method outperformed the previous
methods. In Zh-En data, our method achieved an F1 score of 86.7, which
is 13.3 points higher than that of DiscAlign 73.4 reported in 
\cite{Stengel-Eskin_etal_EMNLP2019}, which is the state-of-the-art
supervised word alignment method.
\citet{Stengel-Eskin_etal_EMNLP2019} used 4M bitexts for pretraining,
while our method needs no bitexts for pretraining. In Ja-En data, our
method achieved an F1 score of 77.7, which is 20 points higher than
that of GIZA++ 57.8 reported in the document attached to
KFTT word alignment \cite{Neubig_KFTT2011}.


For the De-EN, Ro-EN, and En-Fr datasets, \citet{Garg_etal_EMNLP2019},
which is the state-of-the-art unsupervised method, only reported AER in
their paper. For reference, we show the precision, recall, and AER
(based on \cite{Och_Ney_CL2003}) of MGIZA to the same
datasets, as reported in \cite{Zenkel_etal_arXiv2019}\footnote{We took
  these numbers from their GitHub.}.

For the three datasets, we trained our model without distinguishing
between sure and possible and predicted spans without the
distinction. We report both the ordinary measures of precision,
recall, and F1, as well as \citet{Och_Ney_CL2003}'s definition of
precision, recall. We used the scoring script provided by
\citet{Zenkel_etal_arXiv2019} for the latter.

For the De-En and Ro-En datasets, the AERs of the proposed method were 11.4
and 12.2, which are significantly smaller than those of
\cite{Garg_etal_EMNLP2019}: 16.0 and 23.1. For En-Fr,
our method's AER is 9.4, which is significantly larger than the 4.6,
of \cite{Garg_etal_EMNLP2019}. However, if we train our model
using sure alignments and predicts only sure alignments, the AER
of our method becomes 4.0, which is 0.6 smaller than that of
\cite{Garg_etal_EMNLP2019}.






\section{Analysis}

\subsection{Symmetrization Heuristics}

\begin{table}[tb]
  \centering
  \begin{tabular}{ll|rr|r}
    Test set & Method & P & R & F1 \\
    \hline
    Zh-En
         & Zh to En & 89.9 & 85.8 & \textbf{87.8} \\
         & En to Zh & 82.0 & 81.8 & 81.9 \\
         & intersection & 95.5 & 74.9 & 83.9 \\
         & union & 79.4 & 92.7 & 85.5 \\
         & bidi sum th & 84.4 & 89.2 & 86.7 \\
    \hline
    Ja-En
         & Ja to En & 80.6 & 79.7 & \textbf{80.2} \\
         & En to Ja & 61.9 & 69.0 & 65.2 \\
         & intersection & 90.8 & 63.1 & 74.5 \\
         & union & 60.8 & 85.6 & 71.1\\
         & bidi sum th & 77.3 & 78.0 & 77.6 \\
    \hline
    De-En
         & De to En & 89.9 & 85.8 & \textbf{87.8} \\
         & En to De & 82.0 & 81.8 & 81.9 \\
         & intersection & 95.5 & 74.9 & 83.9 \\
         & union & 79.4 & 92.7 & 85.5 \\
         & bidi sum th & 84.4 & 89.2 & 86.7 \\
    \hline
    Ro-En
         & Ro to En & 84.6 & 86.5 & 85.5 \\
         & En to Ro & 87.2 & 86.3 & 86.7 \\
         & intersection & 93.1 & 82.2 & 87.3 \\
         & union & 80.2 & 90.6 & 85.0 \\
         & bidi sum th& 90.4 & 85.3 & \textbf{87.8} \\
    \hline
    En-Fr
         & En to Fr & 79.9 & 91.7 & 85.4 \\
         & Fr to En & 79.5 & 91.3 & 85.0 \\
         & intersection & 85.3 & 88.1 & 86.7 \\
         & union & 75.2 & 94.9 & 83.9 \\
         & bidi sum th & 79.6 & 93.9 & \textbf{86.2} \\
  \end{tabular}
  \caption{Effects of symmetrization for various language pairs}
  \label{tab:symmetrization}
\end{table}

To show the effectiveness of our proposed symmetrization heuristics,
Table~\ref{tab:symmetrization} shows the word alignment accuracies of
two directions, intersection, unison, and the symmetrization method of
Eq.~\ref{eq:symmetrization} , which averages the probabilities of the
two directional predictions and applies thresholding\footnote{We use
  ``bidi sum th'' for the shorthand for this method.}.

For languages whose words are not delimited by white spaces, such as
Chinese and Japanese, the span prediction accuracy ``to English'' is
significantly higher than that of ``from English''. German has the
same tendency because compound words don't have spaces between their
elemental words.  By contract, for languages with spaces between
words, such as Romanian and French, no significant differences exist
between the ``to English'' and ``from English'' accuracies. Since the
proposed symmetrization method of Eq.~\ref{eq:symmetrization} works
relatively well for both cases, we used the heuristics as a default
method for combining the predictions of both directions.

\subsection{Importance of Source Context}

\begin{table}[tb]
  \centering
  \begin{tabular}{ll|rr|r}
    Test set & Context & P & R & F1 \\
    \hline
    Ja-En & no context & 67.3 & 53.0 & 59.3 \\
             & $\pm 2$ words & 73.9 & 70.2 & 72.0 \\
             & whole sentence & 77.3 & 78.0 & 77.6 \\
  \end{tabular}
  \caption{Importance of source context}
  \label{tab:source_context}
\end{table}

Table~\ref{tab:source_context} shows the word alignment accuracies for
questions of different source contexts. Here we used Ja-En data and
found that the source context information is critical for predicting
the target span. Without it, the F1 score of the proposed method is
59.3, which is slightly higher than that of GIZA++, 57.6. If we add
the short context, namely, the two preceding words and the two
following words, the F1 score is improved more than 10 points to
72.0. If we use the whole source sentence as the context, the F1 score
is improved by 5.6 points to 77.6.

\subsection{Learning Curve}

\begin{table}[tb]
  \centering
  \begin{tabular}{lr|rr|r}
    Test set & \# train & P & R & F1 \\
    \hline
    Zh-En  & 300 & 80.9 & 78.4 & 79.6 \\
             & 600 & 82.9 & 81.7 & 82.3 \\
             & 1200 & 82.8 & 85.6 & 84.1 \\
             & 2400 & 83.6 & 87.4 & 85.5 \\
             & 4879 & 84.4 & 89.2 & 86.7 \\
  \end{tabular}
  \caption{Test set performance when trained on subsamples of the
    Chinese gold word alignment data}
  \label{tab:learning_curve}
\end{table}

Table\ref{tab:learning_curve} shows the learning curve of the proposed
method using the Zh-En data. Compared to previous methods, our method
achieves higher accuracy using less training data. Even for
300 sentences, the F1 score of our method was 79.6, which is 6.2 points
higher than that of \cite{Stengel-Eskin_etal_EMNLP2019} (73.4), which
used more than 4800 sentences for training.

It is relatively easy to make gold word alignment of 300 sentences,
once the guideline of the word alignment is established. Since the
proposed method does not require bitexts for pretraining, we assume
that we can achieve a higher word alignment accuracy for low-resource
language pairs than that is currently achieved for high-resource
language pairs using the GIZA++.

\subsection{Zero-shot Word Alignment}

\begin{table}[tb]
  \centering
  \begin{tabular}{lll|rr|r}
    Training & Model & Test & P & R & F1 \\
    \hline
    Ja-En & GIZA++
                         & Ja-En & 59.5 & 55.9 & 57.6 \\
    Zh-En  & Ours 
                         & Ja-En & 55.6 & 62.4 & \textbf{58.8} \\
    \hline
    Ro-En & MGIZA
                         & Ro-En & 90.9 & 61.8 & 73.6 \\
    De-En & Ours
                         & Ro-En & 86.2 & 70.8 & \textbf{77.8} \\
  \end{tabular}
  \caption{Zero-shot word alignment}
  \label{tab:zero_shot}
\end{table}

Since we achieved a higher word alignment accuracy than GIZA++ with as
few as 300 sentences, we tested whether we can perform word alignment
without using the gold alignment of specific language pairs.  Here we
define ``zero-shot word alignment'' as testing the word alignment for
a language pair that is different from the language pair used for
training the model.

Table~\ref{tab:zero_shot} shows the zero-shot word alignment
accuracies. Compared with Table~\ref{tab:comparison}, if we train the
model using Zh-En data, and test it using the Ja-En data, it achieves
an F1 score of 58.8, which is slightly higher than that of GIZA++
(57.6) trained using the Ja-EN data. If we train the model using De-En
data and test it using Ro-En data, it achieves a 77.8 F1, which is 4.2
points higher than that of MGIZA (73.6).

We suspect the reason for the lower zero-shot accuracy between Zh-En
and Ja-En is the difference of tokenization. Although Chinese and
Japanese do share some Chinese characters, Chinese data are character
tokenized, but Japanese data are word tokenized. Future work will seek
a method to utilize word alignment data with different tokenizations.

\section{Related Works}

Although the accuracy of machine translation was improved greatly by
neural networks, the accuracy of word alignment using them cannot
outperform using statistical methods, such as the IBM model
\cite{Brown_etal_CL1993} and the HMM alignment model
\cite{Vogel_etal_COLING1996}.

Recently, \citet{Stengel-Eskin_etal_EMNLP2019} proposed a supervised
method using a small number of annotated data (1.7K-5K sentences) and
significantly outperformed the accuracy of GIZA++.  In this method,
they first mapped the source and target word representations obtained
from the encoder and decoder of the Transformer to a shared space by
using a three-layer feed-forward neural network. They then applied
3 $\times$ 3 convolution and softmax to obtain the alignment
score of the source word and target words. They used 4M parallel
sentences to pretrain the Transformer. We achieved significantly
better word alignment accuracy than
\cite{Stengel-Eskin_etal_EMNLP2019} with less annotated training
data without using parallel sentences for pretraining.

\citet{Garg_etal_EMNLP2019} proposed an unsupervised method that
jointly optimizes translation and alignment objectives. They achieved
a significantly better alignment error rate (AER) than GIZA++ when
they supervised their model using the alignments obtained from GIZA++.
Their model requires about a million parallel sentences for training
the underlying Transformer. We experimentally showed that we can
outperform their results with just 150 to 300 annotated sentences for
training.  We also showed that we can achieve word alignment accuracy
comparable to or slightly better than GIZA++ without using gold
alignments for specific language pairs (zero-shot word alignment).

\citet{Ouyang_McKeown_ACL2019} proposed a monolingual phrase alignment
method using pointer network \cite{Vinyals_etal_NeurIPS2015} that can
align phrases of arbitrary length.  They first segment the source and
target sentences into chunks and compute an embedding for each
chunk. They then use a pointer-network to calculate alignment scores for
each pair of source and target chunks. Compared to our span prediction
method, their method is not flexible because they operate on fixed
target chunks for all fixed source chunks, while our method can change
the target span for each source token.

\section{Conclusion}

We presented a novel supervised word alignment method using
multilingual BERT, which requires as few as 300 training sentences to
outperform previous supervised and unsupervised methods.  We also show
that the zero-shot word alignment accuracy of our method is comparable
to or better than that of statistical methods such as GIZA++.

The future works include utilizing parallel texts in our model. One of
the obvious options is to use XLM \cite{Lample_Conneau_arXiv2019},
which is pretrained on parallel texts, as a drop-in replacement for
multilingual BERT. It is also important to utilize word alignment data
with different tokenizations for languages whose words are not
delimited by spaces.


\bibliography{word_alignment}
\bibliographystyle{acl_natbib}

\end{document}